\documentclass[conference]{IEEEtran}
\IEEEoverridecommandlockouts
\usepackage{arydshln}
\usepackage{cite}
\usepackage{amsmath,amssymb,amsfonts}
\usepackage{algorithmic}
\usepackage{booktabs}
\usepackage{graphicx}
\usepackage{hyperref}
\usepackage{textcomp}
\usepackage[table]{xcolor}
\usepackage{orcidlink}
\usepackage{subcaption}

\def\BibTeX{{\rm B\kern-.05em{\sc i\kern-.025em b}\kern-.08em
    T\kern-.1667em\lower.7ex\hbox{E}\kern-.125emX}}
\begin{document}

\title{Is CLIP the main roadblock for fine-grained open-world perception?%
\thanks{
This work was partially supported by the following projects:
SUN -- Social and hUman ceNtered XR (Horizon Europe R\&I GA:101092612),
FAIR -- Future Artificial Intelligence Research - Spoke 1 (EU NextGenerationEU PNRR M4C2 PE00000013),
ITSERR -- ITalian Strengthening of the Esfri Ri Resilience (EU NextGenerationEU CUP:B53C22001770006).
MUCES -- a MUltimedia platform for Content Enrichment and Search in audiovisual archives (EU NextGenerationEU - PRIN 2022 PNRR P2022BW7CW - CUP: B53D23026090001)}
}

\author{
    \IEEEauthorblockN{
        Lorenzo Bianchi\IEEEauthorrefmark{1}\IEEEauthorrefmark{2}\,\orcidlink{0009-0005-3519-2991},
        Fabio Carrara\IEEEauthorrefmark{1}\,\orcidlink{0000-0001-5014-5089},
        Nicola Messina\IEEEauthorrefmark{1}\,\orcidlink{0000-0003-3011-2487},
        Fabrizio Falchi\IEEEauthorrefmark{1}\,\orcidlink{0000-0001-6258-5313}
    }
    \IEEEauthorblockA{
        \IEEEauthorrefmark{1}\textit{CNR-ISTI}, Pisa, Italy \qquad
        \IEEEauthorrefmark{2} \textit{University of Pisa}, Italy \\
        Email: $<$name$>$.$<$surname$>$@isti.cnr.it
    }
}

\maketitle

\begin{abstract}
Modern applications increasingly demand flexible computer vision models that adapt to novel concepts not encountered during training. This necessity is pivotal in emerging domains like extended reality, robotics, and autonomous driving, which require the ability to respond to open-world stimuli. %
A key ingredient is the ability to identify objects based on free-form textual queries defined at inference time -- a task known as \emph{open-vocabulary object detection}.
Multimodal backbones like CLIP are the main enabling technology for current open-world perception solutions.
Despite performing well on generic queries, recent studies highlighted limitations on the \emph{fine-grained} recognition capabilities in open-vocabulary settings -- i.e., for distinguishing subtle object features like color, shape, and material.
In this paper, we perform a detailed examination of these open-vocabulary object recognition limitations to find the root cause. %
We evaluate the performance of CLIP, the most commonly used vision-language backbone, against a fine-grained object-matching benchmark, revealing interesting analogies between the limitations of open-vocabulary object detectors and their backbones.
Experiments suggest that the lack of fine-grained understanding %
is caused by the poor separability of object characteristics in the CLIP latent space.
Therefore, we try to understand whether fine-grained knowledge is present in CLIP embeddings but not exploited at inference time due, for example, to the unsuitability of the cosine similarity matching function, which may discard important object characteristics. Our preliminary experiments show that simple CLIP latent-space re-projections help separate fine-grained concepts, paving the way towards the development of backbones inherently able to process fine-grained details.  %
The code for reproducing these experiments is available at \url{https://github.com/lorebianchi98/FG-CLIP}.
\end{abstract}

\section{Introduction}

Nowadays, pivotal technologies such as extended reality, autonomous driving, and robotics necessitate more than adherence to closed-set assumptions; they demand the ability to adapt to novel concepts not encountered during the training phase.
Open-vocabulary object detection (OVD) stands out as a critical task for achieving this adaptability, as underlined by several open challenges on egocentric data~\cite{grauman2022ego4d,VISOR2022}.
It involves recognizing objects not included in the training dataset, thus overcoming the limitations inherent in traditional detectors confined to a predefined set of objects.

This flexibility is typically achieved by common-space multi-modal models.
These models embed images and texts in a shared latent space, thanks to the contrastive pre-training performed on large datasets of image-text pairs scraped from the web.
This simplifies similarity calculations, which can be achieved through an efficient dot product between the representations. 
CLIP\cite{radford2021learning} stands out as the most widely utilized model in this category.
Its capabilities become crucial in open-vocabulary object detection, where models typically perform the task by i) detecting regions of the image that are likely to contain objects, ii) computing the similarity between the embedding of the detected image region and that of a set of free-form texts defined at test time, called \textit{vocabulary}.

\begin{figure}[t]
\centering
\begin{subfigure}{\linewidth}
\includegraphics[width=\linewidth,page=1]{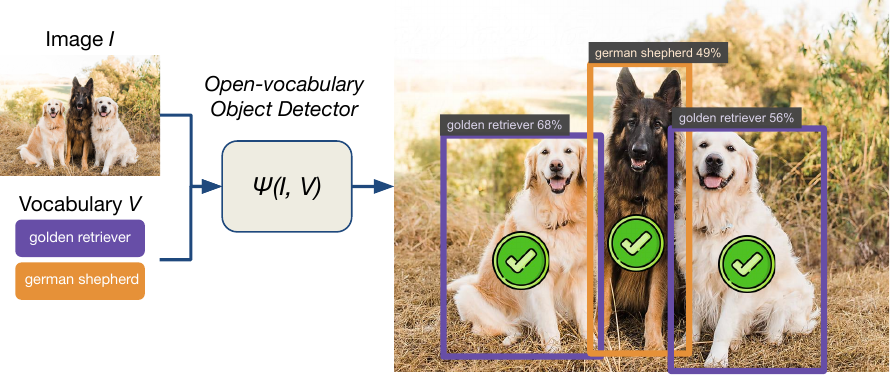}
\caption{Open-vocabulary object detection (OVD)}
\label{fig:ovd_vs_fg-ovd:ovd}
\end{subfigure}
\begin{subfigure}{\linewidth}
\includegraphics[width=\linewidth,page=2]{images/OVD_vs_FG-OVD.pdf}
\caption{Fine-Grained Open-Vocabulary object Detection (FG-OVD)}
\label{fig:ovd_vs_fg-ovd:fg-ovd}
\end{subfigure}
\caption{\textbf{OVD (a) and FG-OVD (b):} in the latter, fine-grained details about the categories to detect are given as free-form text in the input vocabulary.}
\label{fig:ovd_vs_fg-ovd}
\end{figure}

While open-vocabulary detectors excel in generalizing to new concepts not encountered during training, recent studies indicate limitations in capturing fine-grained properties of objects\cite{bianchi2023devil}.
For instance, they may encounter difficulties in distinguishing between a \textit{light brown} dog and a \textit{dark brown} one (\autoref{fig:ovd_vs_fg-ovd}).
One potential explanation for these shortcomings is that CLIP representations may exhibit bias towards category-level concepts while overlooking attribute-level nuances~\cite{chen2023ovarnet}.
To the best of our knowledge, it is not well studied in the literature whether fine-grained properties are absent in the latent space or if these characteristics exist, but trivial matching methods (e.g., dot product, cosine similarity) are insufficient to extract this information.

In this work, we assess the performance of CLIP, the most used backbone in open-vocabulary object detectors, on a fine-grained benchmark to scrutinize its capacity to accurately discern intricate properties of objects.
The findings reveal that the performance of CLIP in fine-grained understanding mirrors that of an open-vocabulary object detector based on CLIP, indicating shared challenges.
This suggests that the fine-grained issues observed in the open-vocabulary object detector may be attributed to the image-text alignment carried out by CLIP rather than to detector localization failures.

Subsequently, by incorporating additional layers on top of frozen visual and textual CLIP encoders and training them on a fine-grained dataset, we illustrate the model's proficiency in accurately assigning the relevant attribute to the object. This shows that the original CLIP embeddings contain fine-grained information that is ignored during the matching phase.

In summary, our paper contributes to the field in the following ways:
\begin{itemize}
    \item We comprehensively evaluate and analyze CLIP's performance on a fine-grained open-vocabulary object detection benchmark. This investigation sheds light on the possibility that challenges faced by open-vocabulary object detectors may be attributed to issues within the CLIP latent space.
    \item We showcase the existence of fine-grained information within CLIP's latent space through the implementation of lightweight architectures trained on frozen CLIP embeddings. This demonstration is substantiated by the model's ability to perform fine-grained matching successfully.
\end{itemize}

\section{Related Work}

\subsection{Image-Text matching}
In recent years, the focus of researchers on image-text matching increased. 
The foundation of the shared space approach for cross-modal matching begins with the exploration of hinge-based triplet ranking loss with hard-negative mining.
This was first attempted with GRU as text extractor \cite{faghri2018vse++}, and later with Transformer Encoder \cite{li2019visual, messina2022aladin, messina2021fine, messina2021transformer, qu2020context, stefanini2021novel, wen2020learning}.

With the growing strength of Transformers also in vision tasks \cite{dosovitskiy2020vit}, many works exploited an early-fusion approach, leveraging Transformers encoder to jointly process images and texts from the very beginning of the proposed architectures\cite{beit3, li2021align, li2020oscar, lu2019vilbert, Su2020VL-BERT, zhang2021vinvl}.
These methods treat image-text matching as a binary classification problem, where, given as input an image-text pair, they train the Transformer architecture to predict the probability that the text correctly describes the image.
While achieving good performance, these models cannot be used in many real-case scenarios since they are computationally expensive at inference time, as they require processing every image-text pair to obtain the score on the whole test set.
Consequently, many methods preferred to exploit a late-fusion approach, keeping the visual and textual pipeline separated\cite{messina2022aladin, messina2021fine, messina2021transformer, sarafianos2019adversarial, wen2020learning, radford2021learning, jia2021scaling}.
This allows separate pipelines for producing the images and text representations, and the similarity score can be computed in a second stage with a simple dot product.

Among these models, CLIP \cite{radford2021learning} stands out as one of the most widely used for performing image-text matching.
The majority of open-vocabulary object detectors rely on the knowledge acquired by CLIP to embed object regions and vocabulary entries in the same feature space \cite{arandjelovic2023three, gu2021open, du2022learning, minderer2022simple, minderer2024scaling, wu2023cora, zhong2022regionclip, zhou2022detecting}.
This approach circumvents the need to conduct inference for each pair of detected image regions and vocabulary entries.

\subsection{Fine-grained understanding}

Although CLIP shows strong performance in tasks such as classification and coarse-grained retrieval, it has shortcomings in associating nuanced properties between sentences and images~\cite{paiss2023teaching, parcalabescu2021valse, ranasinghe2023perceptual, zhong2022regionclip}.
Yuksekgonul et al. \cite{yuksekgonul2022and} suggest that contrastive pretraining does not optimize the model's understanding of the relationships between objects and their attributes.
Furthermore, standard retrieval benchmarks are considered inadequate for assessing the compositional understanding of such models.

Krojer et al. \cite{krojer-etal-2022-image} highlight that vision-language models tend to overlook fine-grained visual information.
Similarly, Chen et al. \cite{chen2023ovarnet} show that representations learned from contrastive pretraining in common-space multimodal models are biased toward category-level concepts rather than attributes, making attribute recognition difficult.

These limitations are even more pronounced in tasks where models usually rely on CLIP latent space to work in an open-world environment, such as open-vocabulary object detection. 
Some recent advanced benchmarks reveal the weaknesses of these models~\cite{bianchi2023devil, bravo2023open}.
These benchmarks not only test the models' ability to generalize to unseen objects during training but also assess their ability to recognize object attributes.
These benchmarks show that current methods are far from achieving satisfactory results.
Our work follows up on those recent findings and investigates potential causes for the performance gap in fine-grained settings.

\section{Method}

\begin{figure*}[t]
    \centering
    \includegraphics[width=\linewidth]{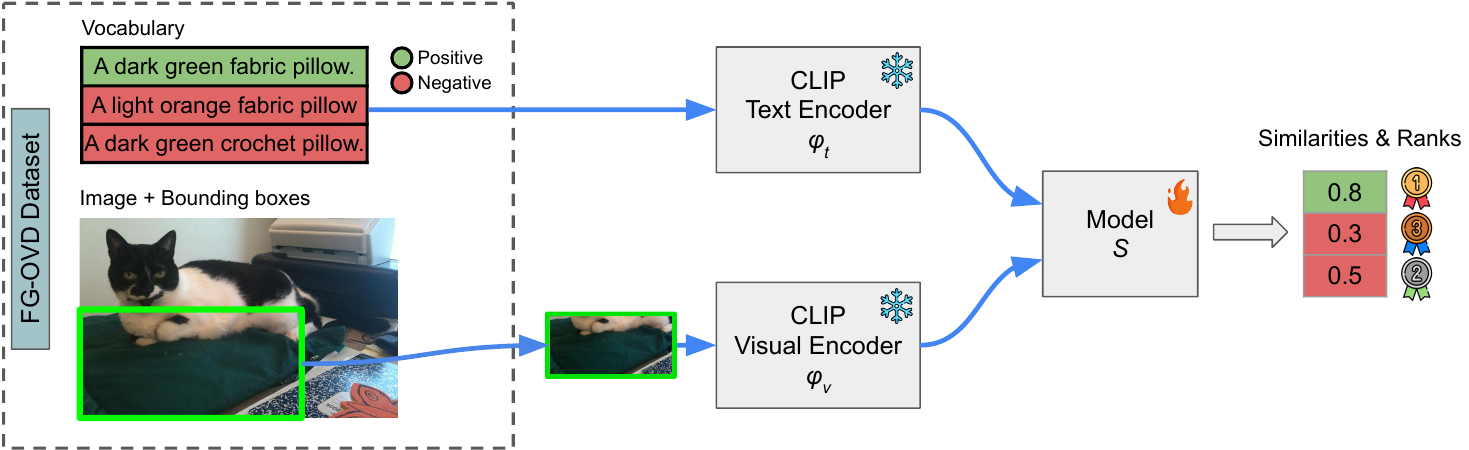}
    \caption{\textbf{FG-OVD Dataset for CLIP Matching.} We leverage the Fine-Grained Open-Vocabulary object Detection (\textbf{FG-OVD}) benchmark suite and training set to investigate our two research questions \textbf{Q1} and \textbf{Q2}. For each object, we extract the corresponding bounding box crop and compute its visual encoding using CLIP. Text embeddings are then generated for the assigned vocabulary entries  (composed by positive + negative captions) associated with the object. Finally, we calculate the similarity and rank of the positive caption between the image crop and the vocabulary entries. To address \textbf{Q1} we use a cosine similarity as model \textit{S}, and the entire pipeline is used only during inference. To address \textbf{Q2}, we choose model S from the solutions described in \autoref{ssec:matching_approaches} and train it on the FG-OVD training set.}

    \label{fig:clip_fg_evaluation}
\end{figure*}

\noindent
In this work, we aim to address two pivotal questions:\\[1ex]
\textbf{Q1.} \emph{Can the limitations observed in open-vocabulary object detection regarding fine-grained understanding be traced back to deficiencies within the CLIP latent space?}\\[1ex]
\noindent
\textbf{Q2.} If yes, \emph{do these limitations arise from an absence of such information in the latent space itself, or is it a consequence of the inadequacy of trivial matching methods (e.g., dot product, cosine similarity) to extract this nuanced information?}\\[1ex]
In this section, we outline the methodologies employed to address these inquiries.

\subsection{CLIP fine-grained evaluation}
\label{sec:fine-grained-evaluation}
Our objective in this part is to address question \textbf{Q1}, aiming to discern whether the failure of open-vocabulary object detectors can be attributed to the image-text matching in the embedding space or to the localization phase. 

To assess the fine-grained knowledge contained in the CLIP latent space and to facilitate a comprehensive analysis in comparison to open-vocabulary object detectors, we used the benchmark suite tailored for Fine-Grained Open-Vocabulary Detection (FG-OVD) proposed in~\cite{bianchi2023devil}.
These benchmarks provide object bounding boxes, each associated with a detailed natural language caption that provides attributes about the object, referred to as a \textit{positive caption}.
In addition, each positive caption is associated with a set of semantically similar but subtly different \textit{negative captions}.

For the purposes of matching rather than detection, we evaluate each object individually.
This adaptation involved retaining a vocabulary specific to each object and cropping the associated bounding box, sparing the model from performing region proposal and detection.
Then, we compute the similarity between the cropped image embedding and the captions embedding using the cosine similarity. This evaluation pipeline is outlined in \autoref{fig:clip_fg_evaluation}.

To benchmark these results, we compare CLIP with OWL~\cite{minderer2022simple, minderer2024scaling}, an open-vocabulary object detector that relies on CLIP.
The performance gap between the two is only explained by the errors introduced by the region proposal and object localization phases, thus providing an estimate of their contribution to open-vocabulary detection performance.

\renewcommand{\v}{\mathbf{v}}
\renewcommand{\t}{\mathbf{t}}
\subsection{Latent Space Characteristics and Matching Approaches}
\label{ssec:matching_approaches}
This section provides the strategy to answer question \textbf{Q2}. 
Assuming that the fine-grained knowledge is present within the CLIP latent space, we hypothesize that the matching scheme used to compare the representations, i.e., the typical cosine similarity, is insufficient to extract this specific information.
To explore this possibility, our strategy involves learning a customized similarity function $S(\v, \t)$, which takes as input the two embeddings $\v$ and $\t$ obtained from the frozen visual and textual encoders, $\phi_v$ and $\phi_t$ respectively.
By forcing $S$ to recognize nuanced object properties based only on the embedded information, we can state that successful results in this regard mean that the embeddings inherently encode fine-grained knowledge. 
To this aim, we will use two distinct datasets having similar image distribution but different annotations. The first one comprises general image-text pairs, which can be used to train and validate our model on standard coarse-grained category-centric classification. Differently, the second dataset is dedicated to training and evaluating the learned function $S$ for fine-grained understanding and is organized in positive and negative captions for each object as explained in Section \ref{sec:fine-grained-evaluation}.

Therefore, the overall training strategy is composed of two steps.
First, we perform a warm-up phase in which we train $S$ on the coarse-grained image-text pairing dataset.%
Then, we fine-tune $S$ using the fine-grained matching dataset.
The first warm-up phase works as a fine-tuning of the original CLIP model, which repurposes the CLIP features to work well with our image distribution and consequently initializes the $S$ function.
This creates a strong and reliable baseline that we can employ as a reference to track the subsequent decline in coarse-grained performance after the fine-tuning step.

We perform the warm-up on the coarse-grained dataset using the \textit{hinge-based triplet loss} as a loss function.
Namely, we optimize
\begin{equation}
\mathcal{L} = \sum_{\substack{i, j \in \mathcal{B}\\i \neq j}} [\alpha + S(\v_i, \t_j) - S(\v_i, \t_i)]_+ + [\alpha + S(\v_j, \t_i) - S(\v_i, \t_i)]_+\,,
\end{equation}
where $S$ is the similarity function, $\mathcal{B}=\{1 \dots B\}$ is a batch, $i$ is the index of an image $I_i$ described by the text $T_i$, while $j \neq i$ is the index of a negative image $I_j$ and a negative text $T_j$ taken from the batch.
$\phi_v$ and $\phi_t$ are respectively the visual and textual CLIP encoders, and $\v_i = \phi_v(I_i)$ and $\t_i = \phi_t(T_i)$ are the frozen encoded text and image, respectively.
$\alpha$ is the margin of separation.

We then perform training on the fine-grained dataset, where each image $I_i$ is associated with a vocabulary composed of a positive caption and a set of $N$ similar but slightly different captions, called negative captions.
Again, we rely on the \textit{hinge-based triplet loss}, but this time, we adapt it for the fine-grained discrimination task: for each tuple (image, vocabulary), we take as anchor the image, as positive the correct caption, and as negative the incorrect ones.
Formally,

\begin{equation}
\mathcal{L}_\text{FG} = \sum_{i=1}^B \sum_{j=1}^N [\alpha + S(\v_i, \t_{i,j}^\text{neg}) - S(\v_i, \t_i^\text{pos})]_+\,,
\end{equation}
where $B$ is the batch size, $\v_i$ is the encoding of the $i$-th image, $\t_i^\text{pos}$ is the encoding of the positive caption associated with the $i$-th image, $\t_{i,j}^\text{neg}$ is the encoding of the $j$-th negative caption associated with image $i$, $\alpha$ is the minimum separation margin that should hold between positive and negative captions, and $[x]_+ \equiv \max(x, 0)$.

We evaluate different implementations of the matching function $S$: %
\subsubsection{Baseline (CLIP matching function)}
\begin{equation}\label{eq:baseline}
    S(\v,\t) = \cos(\v, \t)
\end{equation}
The vanilla cosine similarity represents the commonly used matching function used in CLIP and open-vocabulary object detectors to match visual and textual representations.

\subsubsection{Linear projection layer}
\begin{equation}\label{eq:linear}
    S(\v,\t) = \cos(W_v\v+b_v, W_t\t+b_t)
\end{equation}
We propose two linear projection layers that operate on top of the frozen visual ($\v$) and textual ($\t$) features, aiming to project the embeddings into a space of the same dimensionality.
The final matching function is kept as the cosine similarity between the transformed feature vectors.
We aim to explore the feasibility of linearly separating fine-grained concepts (if they exist) embedded in the CLIP embeddings.

\subsubsection{Linear projection layer only above text encoder}
\begin{equation}\label{eq:linear:text-only}
    S(\v,\t) = \cos(\v, W_t\t+b_t)
\end{equation}
We introduce a linear projection layer solely above the text encoder while keeping the visual embedding fixed. This setup forces the image latent space to maintain coherence with the original CLIP space while repurposing only textual representations.

\subsubsection{Linear projection layer only above visual encoder}
\begin{equation}\label{eq:linear:visual-only}
    S(\v,\t) = \cos(W_v\v+b_v, \t)
\end{equation}
Similarly, we apply a linear projection layer solely above the visual encoder, following the same principle as the previous scheme.
By projecting only one embedding modality at a time, we can explore any potential asymmetry in the results and gain insights into the role of each modality in capturing fine-grained information.

\subsubsection{MLPs layer}
\begin{equation}\label{eq:mlp}
    S(\v,\t) = \cos(\text{MLP}_v(\v), \text{MLP}_t(\t))
\end{equation}

We incorporate two Multi-Layer Perceptrons (MLPs) with a non-linear activation function to repurpose the embeddings before performing the cosine similarity. This scheme aims to explore the effects of non-linearity on both coarse-grained and fine-grained performance and to compare these results with the effects observed with linear projection approaches.

\subsubsection{Attention layer}
\begin{equation}\label{eq:mha}
    S(\v,\t) = \sigma \left ( \text{MHA}([\text{CLS}, \v, \t])^{(0,0)} \right )
\end{equation}
We build a multi-head attention layer above the two encoders to compute self-attention utilizing the two embeddings along with a to-be-learned CLS token of the same dimensionality. Then, after the attention layer, we apply the sigmoid function to the first element of the CLS token output. The score of the sigmoid represents the similarity between the two embeddings. This approach explores a more complex and expressive non-linear alternative to the MLP that exploits attention to automatically weight the contribution of image and text features.

\section{Experiments}

\subsection{Dataset and Metrics}

\begin{figure*}[h!]
    \centering
    \includegraphics[width=\linewidth]{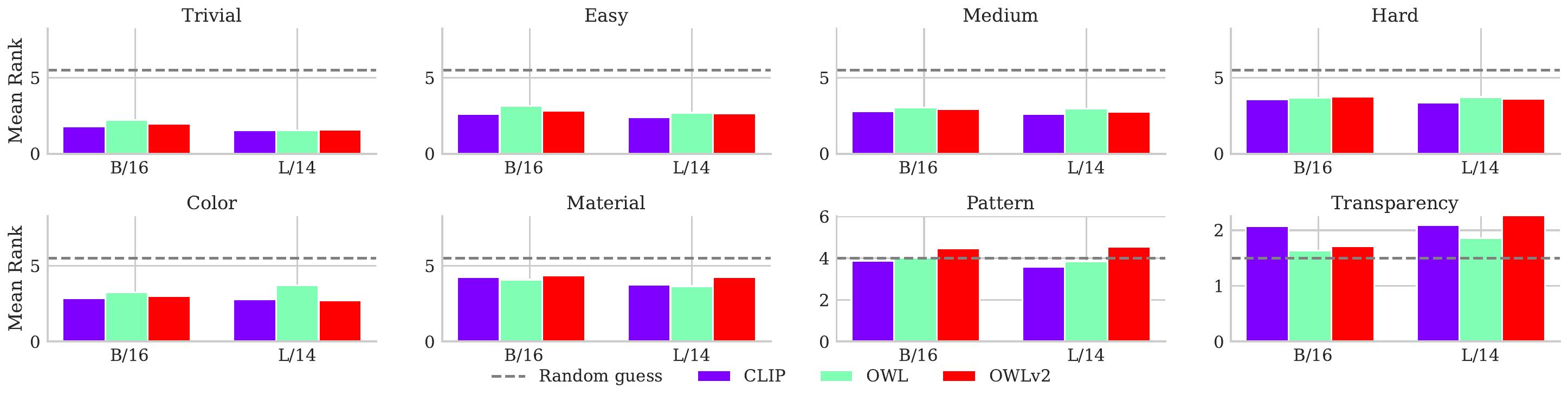}
    \caption{\textbf{CLIP vs. OWL in fine-grained understanding.} We evaluate CLIP and OWL, configured as B/16 and L/14, against the Difficulty-based (first row) and Attribute-based (second row) FG-OVD benchmarks. The bar graph shows the Mean Rank of the positive label \textbf{(lower is better)}, which represents the average position assigned by the model to the correct label within the overall vocabulary. Vocabulary lengths vary, with 3 for transparency, 8 for pattern, and 11 for other attributes.}
    \label{fig:clip_vs_owl}
\end{figure*}
\begin{table*}
\centering
\caption{\textbf{Coarse-grained and Fine-grained performance.} We analyze the performance of the investigated similarity functions $S$ after a warm-up train on COCO and a subsequent fine-tuning on the fine-grained dataset FG-OVD. In the fine-tuned configuration (+FG-OVD rows), we denote the delta between these results and those obtained during the warm-up in parentheses.}
\label{tab:results}

\newcommand{\g}[1]{ \textcolor{green}{(#1)}}
\renewcommand{\r}[1]{ \textcolor{red}{(#1)}}
\renewcommand{\b}[1]{\textbf{#1}}
\renewcommand{\u}{$\uparrow$}
\renewcommand{\d}{$\downarrow$}
\newcommand{\m}[2]{\multicolumn{#1}{c}{#2}}

\begin{tabular}{llllllll}
\toprule
              &                  & \m{6}{COCO Retrieval} \\
                                   \cmidrule(lr){3-8}
              & \m{1}{FG-OVD}    & \m{3}{I$\to$T}              & \m{3}{T$\to$I} \\
              \cmidrule(lr){2-2}   \cmidrule(lr){3-5}            \cmidrule(lr){6-8}
              & Mean Rank \d & R@1 \u       & R@5 \u       & R@10 \u      & R@1 \u       & R@5 \u       & R@10 \u      \\
\midrule
\b{CLIP B/16} & 2.98 & 41.5 & 65.9 & 76.2 & 22.6 & 44.1 & 54.9 \\
\midrule
\b{Linear (both)} & 3.78 & \textbf{48.0} & \textbf{76.6} & \textbf{85.4} & \textbf{37.2} & \textbf{65.6} & \textbf{76.6}         \\
\quad +FG-OVD & 1.46\g{-2.32} & 37.1\r{-10.9} & 66.8\r{-9.8} & 78.4\r{-7.0} & 35.6\r{-1.6} & 63.9\r{-1.7} & 75.0\r{-1.6} \\
\midrule
\b{Linear (visual only)} & 3.53 & 45.8 & 74.2 & 83.7 & 35.4 & 64.2 & 75.3         \\
\quad +FG-OVD & 1.54\g{-1.99} & 39.4\r{-6.4} & 69.5\r{-4.7} & 79.8\r{-3.9} & 34.3\r{-1.1} & 62.9\r{-1.3} & 74.1\r{-1.2} \\
\midrule
\b{Linear (text only)} & 3.48 & 47.3 & 75.1 & 84.9 & 36.0 & 64.3 & 75.6         \\
\quad +FG-OVD & 1.57\g{-1.91} & 41.1\r{-6.2} & 70.4\r{-4.7} & 80.6\r{-4.3} & 34.7\r{-1.3} & 63.2\r{-1.1} & 74.6\r{-1.0} \\
\midrule
\b{MLP} & 3.49 & 45.9 & 75.5 & 84.6 & 36.5 & 64.9 & 76.2         \\
\quad +FG-OVD & \textbf{1.43\g{-2.06}} & 31.9\r{-14.0} & 60.2\r{-15.3} & 72.7\r{-11.9} & 33.6\r{-2.9} & 62.0\r{-2.9} & 73.9\r{-2.3} \\
\midrule
\b{MHA} & 4.08 & 36.3 & 66.1 & 78.1 & 29.1 & 57.6 & 70.3         \\
\quad +FG-OVD & 1.54\g{-2.54} & 22.3\r{-14.0} & 48.3\r{-17.8} & 61.2\r{-16.9} & 22.6\r{-6.5} & 49.2\r{-8.4} & 62.0\r{-8.3} \\
\bottomrule
\end{tabular}
\end{table*}
For the warm-up phase and to evaluate the coarse-grained capabilities of our model, we selected the MS-COCO dataset as our image-text pair dataset. We followed the partitioning introduced by Karpathy et al.\cite{karpathy2015deep}, reserving 113,287 images for training, 5,000 for validation, and 5,000 for testing, each with five captions. We evaluated retrieval performance on COCO using the Recall@$k$ metric to measure the ability of our model to retrieve relevant text or images accurately. Specifically, Recall@$k$ assesses the proportion of queries that successfully retrieve the correct item within the first $k$ results. %

We used the Fine-Grained Open-Vocabulary Object Detection (FG-OVD)~\cite{bianchi2023devil} suite for fine-grained training and evaluation, adapting it for classification using crops of object regions instead of detection.
This suite associates each object with a customized vocabulary consisting of a detailed sentence describing the object and its attributes, called the positive caption, and a set of negative captions subtly altering certain attributes (see \autoref{fig:clip_fg_evaluation}).
The benchmarks within the suite are categorized by difficulty level (Trivial, Easy, Medium, and Hard), with the degree of change in the captions decreasing as the selected benchmark becomes more difficult. For example, the Hard benchmark indicates that only one attribute is changed in the negative labels. In addition, the suite includes attribute-based benchmarks (Color, Material, Pattern, and Transparency), where only attributes of a particular type are changed, making it easier to evaluate model performance in each attribute category.
To evaluate on this benchmark, we measured the Mean Rank of the positive label within the vocabulary when ranked by descending matching score. As for the number of negative labels in the vocabulary $N$, we follow \cite{bianchi2023devil} and choose 10 for Trivial, Easy, Medium, Hard, Color and Material, 7 for Pattern, and 2 for Transparency. In \autoref{ssec:q1}, we plot the Mean Rank for each benchmark, while in \autoref{ssec:q2}, we report the mean of the values obtained from the eight benchmarks.

For training, we used the Hard training set with $N=10$ negatives.

\subsection{Implementation details}

For the warm-up phase on COCO, we train the matching function $S$ with the Adam optimizer, using a learning rate of 5e$^{-4}$ for 10 epochs and a triplet loss margin $\alpha=0.2$.
For fine-tuning on the FG-OVD, we use Adam with a reduced learning rate (1e$^{-5}$) and set the triplet loss separation margin $\alpha=0.05$. Maintaining a deliberately low margin is critical, as higher values were observed to disrupt the alignment established during the warm-up phase, significantly reducing retrieval performance on COCO. The fine-tuning process includes 10 epochs.
Regarding the newly added layers, we set up the attention layer with 64 heads, while the MLPs feature 2 layers with 512 neurons each and the $tanh$ activation function.
We train all the proposed architectures using embeddings extracted from CLIP B/16.

\subsection{Impact of object localization is marginal in FG-OVD}
\label{ssec:q1}
\autoref{fig:clip_vs_owl} compares a CLIP-based open-vocabulary object detector with vanilla CLIP applied to pre-segmented image regions on the FG-OVD benchmark suite. 
It is important to note that the task of the detector is more challenging than that of CLIP. Indeed, the detector must also localize the object rather than solely classifying it.
We make the comparison with OWL, an open vocabulary object detector based on CLIP, presented in both its original~\cite{minderer2022simple} and second~\cite{minderer2024scaling} versions, with the same backbones (B/16 and L/14). 

Looking at the results, the performance of CLIP mirrors the pattern shown by the detectors. However, the overall performance of CLIP remains relatively low.
For example, in the Hard benchmark, the model ranks the correct caption on average around 4th out of 11 possible captions.
This highlights the significant limitations of classification using the CLIP latent space and points to the need for significant improvements in open-world fine-grained classification.

Despite facing a more challenging task than CLIP, the detectors' performance is not significantly lower. Their results are consistent with CLIP's patterns, showing similar trends in both the Difficulty-based and Attribute-based benchmarks, with better performance observed in the Color benchmark (an attribute more commonly found in web-scraped images), while the Material and Pattern benchmarks show lower performance. This suggests that the recent challenges open vocabulary object detectors face in fine-grained understanding are more related to classification within the shared image-text embedding space than to the localization phase.

\subsection{A linear projection is enough for fine-grained matching}
\label{ssec:q2}

The results presented in \autoref{tab:results} illustrate the performance of the proposed architecture after the warm-up on COCO and the subsequent fine-tuning on FG-OVD.

Observing the results, the warm-up on COCO improves the performance of COCO retrieval compared to the original CLIP, indicating increased specialization in the characteristics of COCO captions and images. Conversely, performance on the fine-grained benchmark is poor because COCO captions generally lack detailed attributes about objects.

After fine-tuning on the fine-grained dataset, there's a slight decrease in retrieval performance on COCO. However, the key observation is the significant decrease in the FG-OVD Mean Rank. This suggests that the newly added layers, relying solely on information from the embeddings, effectively discriminate the correct attributes for object mapping. %
This addresses our question \textbf{Q2}, where we wondered if fine-grained knowledge was missing in the CLIP latent space or if we were not using the right tools to extract it.
These results show that we can learn a more complex similarity matching between the representations and that nuanced information is indeed present in CLIP embeddings.
In addition, the fact that these results can be achieved only with a linear projection, demonstrate that this type of information can be linearly separated in the embedding space. 

Comparing the MLPs with the linear projection, it is evident that nonlinearity does not provide any advantage in maintaining a favorable trade-off between fine- and coarse-grained performance.
Although the results on the fine-grained benchmarks are slightly better, the retrieval performance on COCO worsens.

The results obtained with a linear projection applied to only one embedding modality reveal interesting observations. During the warm-up phase, these projections demonstrate superior fine-grained performance compared to models with both embeddings repurposed.
This phenomenon can be attributed to the fact that only one projection needs to be learned, which is forced to maintain coherence with the original CLIP latent space. Consequently, this leads to a smaller fine-grained performance drop during the warm-up phase, which is usually due to the limited fine-grained attribute information within the COCO dataset. In addition, this adherence to the original embedding space results in a minor coarse-grained performance degradation after fine-tuning compared to other tested configurations. 
What is most compelling is that despite the fine-grained results being slightly lower compared to the other matching functions, there is a great improvement in the fine-grained results despite the need to adhere to the CLIP latent space. 
This suggests that fine-grained properties are not only present and linearly separable within the CLIP embedding space but can even emerge with a simple linear adjustment of the embedding of only one modality.

Experiments with the multi-head attention layer suggest that more complex and expressive architecture can better adapt to the novel fine-grained task but with a higher risk of overfitting the training data and disrupting the original space.

\section{Conclusion}
We studied the challenges confronting current state-of-the-art cross-modal image-text models in achieving open-world understanding. We began our analysis by examining the relationship between open-vocabulary object detectors and their vision-language backbone, specifically focusing on CLIP. Our results suggest that localization is marginal in the limitations observed in fine-grained open-vocabulary object detection. This demonstrates that the primary problem lies in the interaction between vision and language within the shared latent space.

Furthermore, while fine-grained information exists within the CLIP latent space, the representation is heavily biased towards coarse-grained concepts. This bias causes similar concepts to be positioned too closely within the latent space, making it difficult to detect nuanced differences using traditional cosine similarity. Importantly, we demonstrated that fine-grained information is still linearly separable within latent space despite this bias.

In the future, we aim to explore better pre-training strategies to construct more balanced image-text representations that effectively incorporate fine- and coarse-grained features. In addition, we will investigate alternative matching functions capable of extracting fine-grained features within the CLIP latent space without the need for task-specific datasets to learn this function.

\bibliographystyle{plain} 
\bibliography{bibtex}

\vspace{12pt}

\end{document}